 \documentclass[sigconf]{acmart}
\usepackage{cleveref}
\usepackage{multirow}
\usepackage{enumitem}

\AtBeginDocument{%
  \providecommand\BibTeX{{%
    \normalfont B\kern-0.5em{\scshape i\kern-0.25em b}\kern-0.8em\TeX}}}

\copyrightyear{2023}
\acmYear{2023}
\setcopyright{rightsretained}
\acmConference[FAccT '23]{2023 ACM Conference on Fairness, Accountability, and Transparency}{June 12--15, 2023}{Chicago, IL, USA}
\acmBooktitle{2023 ACM Conference on Fairness, Accountability, and Transparency (FAccT '23), June 12--15, 2023, Chicago, IL, USA}
\acmDOI{10.1145/3593013.3594094}
\acmISBN{979-8-4007-0192-4/23/06}

\begin{document}

\title[Examining risks of racial biases in NLP tools for child protective services]{Examining risks of racial biases in NLP tools \\for child protective services}

\author{Anjalie Field}
\email{afield6@jhu.edu}
\orcid{0000-0002-6955-746X}
\affiliation{%
  \institution{Johns Hopkins University}
  \city{Baltimore}
  \state{MA}
  \country{USA}
}
\additionalaffiliation
{
 \institution{Carnegie Mellon University}
 \city{Pittsburgh}
 \state{PA}
 \country{USA}
}

\author{Amanda Coston}
\email{acoston@andrew.cmu.edu}
\orcid{0000-0001-9282-9921}
\affiliation{
  \institution{Carnegie Mellon University}
  \city{Pittsburgh}
  \state{PA}
  \country{USA}
}

\author{Nupoor Gandhi}
\email{nmgandhi@andrew.cmu.edu}
\orcid{0009-0008-2491-4938}
\affiliation{%
  \institution{Carnegie Mellon University}
  \city{Pittsburgh}
  \state{PA}
  \country{USA}
}

\author{Alexandra Chouldechova}
\email{achoulde@andrew.cmu.edu}
\orcid{0000-0002-2337-9610}
\affiliation
{
 \institution{Carnegie Mellon University}
 \city{Pittsburgh}
 \state{PA}
 \country{USA}
}
\additionalaffiliation{%
  \institution{Microsoft Research NYC}
  \city{New York}
  \state{NY}
  \country{USA}
}

\author{Emily Putnam-Hornstein}
\email{eph@unc.edu}
\orcid{0000-0003-1581-6582}
\affiliation{%
  \institution{The University of North Carolina at Chapel Hill}
    \city{Chapel Hill}
  \state{NC}
  \country{USA}
}

\author{David Steier}
\email{steier@andrew.cmu.edu}
\orcid{0009-0009-4184-6209}
\affiliation{%
  \institution{Carnegie Mellon University}
  \city{Pittsburgh}
  \state{PA}
  \country{USA}
}

\author{Yulia Tsvetkov}
\email{yuliats@cs.washington.edu}
\orcid{0000-0002-4634-7128}
\affiliation{%
  \institution{University of Washington}
    \city{Seattle}
  \state{WA}
  \country{USA}
}

\acmConference[FAccT]{ACM Conference on Fairness, Accountability, and Transparency}{June 12-15}{Chicago, IL}

\renewcommand{\shortauthors}{Field, et al.}

\newcommand{\afield}[1]{}
\begin{abstract}
Although much literature has established the presence of demographic bias in natural language processing (NLP) models, most work relies on curated bias metrics that may not be reflective of real-world applications.
At the same time, practitioners are increasingly using algorithmic tools in high-stakes settings, with particular recent interest in NLP.
In this work, we focus on one such setting: child protective services (CPS).
CPS workers often write copious free-form text notes about families they are working with, and CPS agencies are actively seeking to deploy NLP models to leverage these data.
Given well-established racial bias in this setting, we investigate possible ways deployed NLP is liable to increase racial disparities.
We specifically examine word statistics within notes and algorithmic fairness in risk prediction, coreference resolution, and named entity recognition (NER).
We document consistent algorithmic unfairness in NER models, possible algorithmic unfairness in coreference resolution models, and little evidence of exacerbated racial bias in risk prediction. While there is existing pronounced criticism of risk prediction, our results expose previously undocumented risks of racial bias in realistic information extraction systems, highlighting potential concerns in deploying them, even though they may appear more benign.
Our work serves as a rare realistic examination of NLP algorithmic fairness in a potential deployed setting and a timely investigation of a specific risk associated with deploying NLP in CPS settings.
\end{abstract}




\maketitle

Natural Language Processing (NLP) models are well-known to absorb and amplify data biases.
A plethora of research has shown that models exhibit gender bias \citep{sun-etal-2019-mitigating}, with more recent work also examining dimensions like race \citep{field2021Survey}, disability \citep{hutchinson-etal-2020-social}, and mental health status \citep{lin2022gendered}. 
Despite these findings, understanding of model biases in realistic settings remains limited. Most work focuses on developing benchmark tasks and draws data from laboratory psychology studies or large public records \citep{blodgett-etal-2020-language,field2021Survey}.
Curated data sets facilitate reproducibility and controlled experiments, but they can fail to articulate what is being measured, contain ambiguities, and may not reflect real deployment settings \citep{blodgett-etal-2021-stereotyping}.

At the same time, despite concerns about bias, practitioners are increasingly turning to machine learning in high-stakes settings such as public services, hiring, education, and criminal justice \cite{Saxena2020Human, vaithianathan2017developing, Raghavan2020Mitigating, chouldechova2017fair, studentsatrisk}.
We focus on one such setting: the child protection system (CPS).
CPS agencies have copious free-form text notes written by CPS workers, which contain extensive details, including professional assessments of family situations and needs \cite{Saxena2020Human, guideforcaseworkers}.
Undirected manual reviews of notes is challenging for caseworkers and supervisors when, for example, making time-sensitive decisions or examining a newly assigned case \citep{perron2019detecting}.
Further, needing to spend time on administrative tasks like reviewing notes rather than working directly with families is associated with higher caseworker turnover and worse experiences for affected children \cite{strolin2008should,strolin2010listening}.
 As the potential for NLP to aid in processing expert-written notes has been demonstrated in domains like healthcare and law \citep{ji-etal-2021-medical,johnson2016mimic,uzuner20112010,zhong-etal-2020-nlp}, CPS agencies are actively seeking NLP tools to extract and deliver information from unstructured data \citep{perron2019detecting,victor2021automated,hsu-etal-2020-characterizing,dhsRFP2018},
  and some research has additionally discussed implications of leveraging these data in predictive risk models \citep{Saxena2020Human, moon2023rethinking}.
Predictive risk models have already been implemented to inform various aspects of CPS practice, such as which allegations are screened-in for investigation \citep{chouldechova18a,nash2017examination,Saxena2020Human} or which investigations are prioritized for supervisory review \citep{Putnam-HornsteinLA}, but existing models  primarily rely on tabular structured data \cite{Saxena2020Human}.

In this work, we examine the algorithmic fairness of NLP technology in CPS settings.
Research on NLP model performance over real high stakes data is extremely difficult, given challenges around data privacy and forming partnerships with practitioners.
Unlike research that constructs data sets to probe NLP model bias, our work serves as a rare opportunity to benchmark biases in a realistic setting.
Furthermore, families involved in CPS already express mistrust in ``the system'' \citep{Brown2019} and there is pronounced criticism of using algorithmic tools in any capacity \citep{eubanks2018automating}.
Thus, our work is also a timely exploration of one potential risk of deploying NLP in CPS settings: algorithmic unfairness. 

We focus specifically on disparities in model performance over case notes about black and white families based on decades of research demonstrating racial disparities in the child protection system in the United States \cite{roberts2009shattered, dettlaff2011disentangling, hill2004institutional, wells2009bias, roberts2019digitizing, USReport2017}.
In comparison to white children, black children are disproportionately involved in the system \citep{Cenat2021}, they may be more likely to be reported for abuse by doctors even with similarly severe injuries \citep{lane2002racial}, their referrals are investigated at higher rates \citep{child_maltreatment_2021}, and in some states they are placed in foster care at higher rates \citep{foster_care_2021}.
Institutional racism in child protective services has also been attributed to the links between the child protection system and other systems like mental health services, criminal justice, and education \cite{hill2004institutional}.
We investigate how deployed NLP models may amplify these disparities.

In Section~\ref{sec:cps_data}, we first describe our primary data set: more than 3 million contact notes written by caseworkers, supervisors, and service providers in the Department of Human Services (DHS)
 in one anonymous USA county.\footnote{All research was conducted with IRB approval and under a data sharing and protection agreement with the county.} Next, we present initial statistics surfacing possible racial disparities in the text data (Section~\ref{sec:data_statistics}). We then examine two types of NLP models that could be deployed in this setting: the incorporation of text data into an existing risk predictive tool (Section~\ref{sec:risk_assessment}) and information extraction tools, specifically named entity recognition (NER) and coreference resolution (Section~\ref{sec:ner}). Despite substantial research on gender bias in coreference systems \citep{rudinger-etal-2018-gender,zhao-etal-2018-gender}, to the best of our knowledge, our work is the first consider racial bias.
Our results show consistent racial bias in NER, possible biases in coreference resolution, and no evidence of increased racial bias in risk assessment.
Thus, while risk assessment systems are already heavily scrutinized \citep{moon2023rethinking}, our work highlights one way deployment of information extraction systems could also result in direct harms, even though they are further removed from direct decision-making and may appear more benign than risk prediction.

Finally, we emphasize algorithmic fairness is only one risk associated with developing NLP technology in CPS settings. Our focus on this particular risk is motivated by extensive NLP research on synthetically probing model biases with few examinations of model performance in realistic settings, but broader sociotechincal forces must be considered \citep{blodgett-etal-2020-language}.
Thus, we conclude by discussing some of the risks beyond algorithmic fairness identified in prior work and surfaced in our analysis \citep{moon2023rethinking}.
Our work is a rare documentation of direct harms that can result from algorithmic bias in deployed NLP systems.
To the best of our knowledge, it is the first work to consider algorithmic unfairness in NLP technology for CPS settings.

\section{Data: CPS Contact Notes}
\label{sec:cps_data}
Child welfare cases typically begin with a referral, where someone (the ``reporter'') contacts social services with concerns about a child.
A call-screening staff member then makes a \textit{Call Screen Decision}: whether or not to investigate the allegations made in the referral.
If the referral is \textit{screened in}, a caseworker then conducts an investigation, which may involve interviewing relevant contacts and conducting assessments. In many states, caseworkers must complete the investigation in a fixed amount of time, such as 60 days \cite{@making_child}. Based on the investigation, the caseworker decides whether or not the family should be accepted for services by the child protection agency (\textit{Service Decision}). If the family is accepted, a case is opened. Cases can stay open for varying lengths of time, and families may receive a range of services, such as housing support or addiction treatment often through external \textit{service providers}.
Caseworkers, supervisors, and service providers write copious notes throughout the duration of a case, including during the investigation phase before a case is actually opened.

\begin{figure}[!ht]
    \centering
    \includegraphics[width=0.45\textwidth]{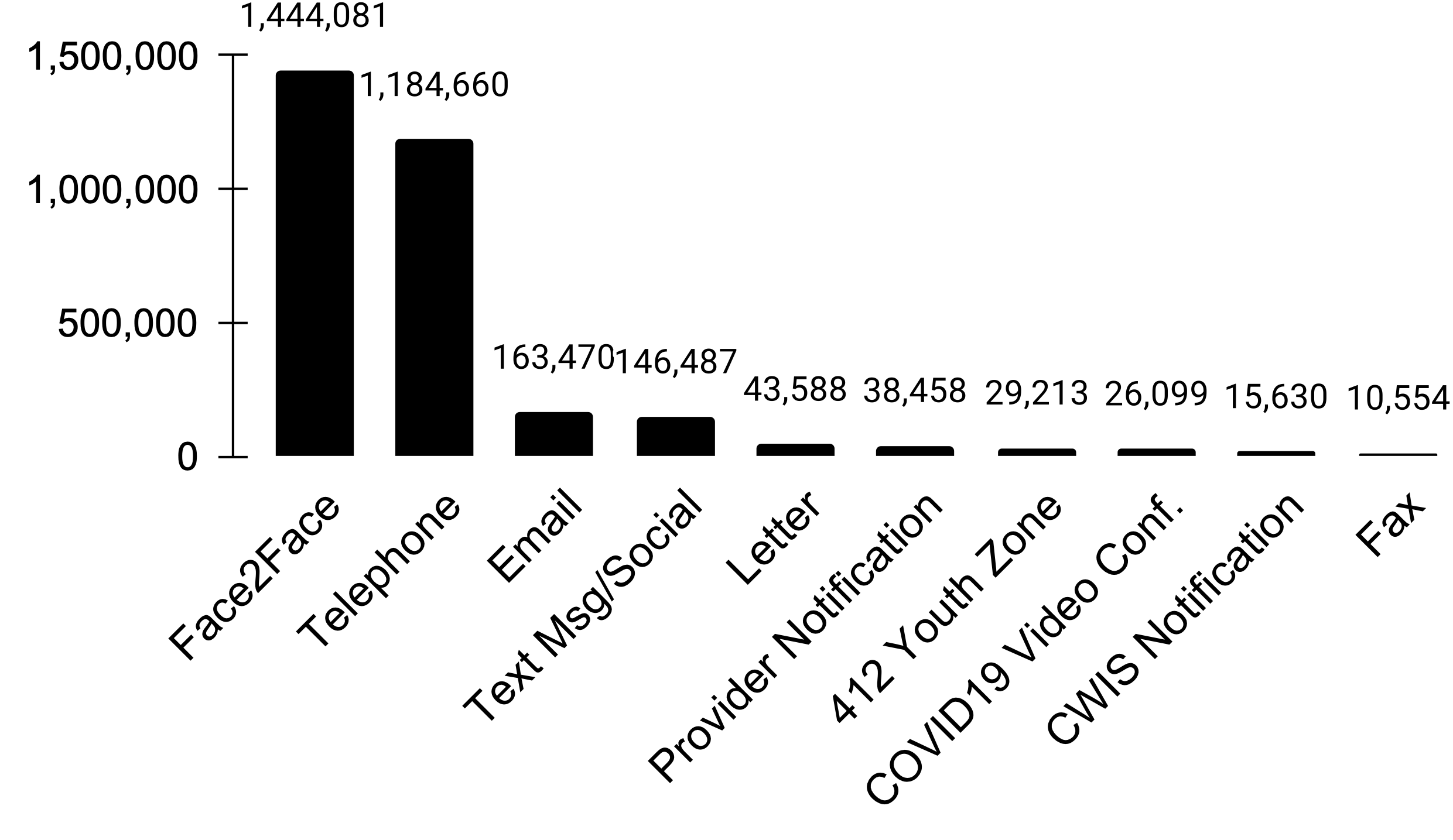}
    \Description{Bar chart histogram, where the x-axis specifies types of contact. The one with the highest bar is Face2Face, followed by Telephone, Email, Text Msg/Social, Letter, Provider Notification, 412 Youth Zone, COVID19 Video Conf., CWIS Notification, and Fax}
\caption{Histogram of contact types for 3.1M contact notes associated with cases and referrals. Contact types with <1000 notes are omitted for brevity.}
\label{fig:contact_type}
\end{figure}

In this work, we investigate a data set of 3,105,071 contact notes, which consists of all \textit{contact notes} written by CPS workers in the Department of Human Services (DHS) in an anonymous county from approximately 2010 to November 23, 2020 (notes prior to 2010 were inconsistently digitized in the current system).
\Cref{tab:overview_stats} presents overview statistics of the data.
Contact notes log communication between CPS workers and families. As shown in \Cref{fig:contact_type}, most notes record telephone or face-to-face contacts, such as visiting families at home or school, but notes can also be created for other forms of contact, such as emails.
In addition to the primary data, we also reference associated meta-data, such as case open and close dates, lists of associated clients, and mappings between cases, referrals, and clients.
Throughout this work, when we refer separately to \textit{referral} notes and \textit{case} notes we generally do not duplicate data, e.g. for a referral that turned into a case, we would not consider notes associated with the referral as also associated with the case, which is consistent with DHS data organization.

\begin{table}
\begin{tabular}{cccc}
        \hline
        & Average & Min. & Max. \\
        \hline
        \hline
         \# tokens per note & 156.59 & 0 & 2,915 \\
         \# notes per case & 128.36 & 1 & 4,831  \\
         \# notes per referral & 8.55 & 1 & 672 \\
         \hline
\end{tabular}
\caption{Overview statistics for 3.1M contact notes associated with cases and referrals.}
\label{tab:overview_stats}
\end{table}

Research was conducted with full-board IRB approval and under a data sharing agreement with DHS. In accordance with these protocols, data was exclusively stored on a remote disk-encrypted server, with access restricted to approved members of the research team who underwent IRB-determined training, and researchers only accessed the server through secure encrypted connections.
In general the standards of security and privacy upheld throughout this research were higher than those mandated for DHS contractors.

\subsection{Conceptualization of Race}
\label{sec:race_definition}

As discussed, we focus on investigating algorithmic racial bias motivated by limited research on algorithmic racial bias in realistic NLP systems and documented disparities in CPS settings. This investigation necessitates information about races of families involved in CPS.
Our primary source of information about race is metadata entered by CPS workers, which means our conceptualiztion of race is race as perceived by CPS workers. This conceptualiztion is appropriate in some contexts, e.g. how implicit bias of CPS workers may manifest in text data, but as perceived race can differ from self-identified race \citep{roth2016multiple}, it suggest our results may not be fully reflective of experiences of families.

We focus specifically on white and black families and clients due to well-document disparities between white and black families in CPS settings and for clarity of analysis.
Although a clear over-simplification, this focus does include much of the available data: out of all 234,818 clients listed on referrals, meta-data specifies 47.3\% as ``White'', 33.7\% as ``Black or African American'', and and 11.1\% as ``Unknown/Did Not Ask/Declined to Answer'', with <8\% of the data specifying client race as mixed or other than white or black.
In general, rerunning statistics (e.g., Section~\ref{sec:data_statistics}) where we include other/mixed race clients as either black or white does not change results.
Nevertheless, this simplification is a limitation in our work.
We provide additional details on exact processing of race information for each task in task-specific sections (Section~\ref{sec:data_statistics_methods}, \ref{sec:ner_methods}).

\section{Racial Disparities in CPS Case Notes}
\label{sec:data_statistics}

We first consider possible racial disparities in the raw text data, which could have implications for both training and deploying NLP models in this setting.
For example, if there tend to be more and longer notes written about white families than black families, any models \textit{trained} on this data may learn patterns disproportionally representative of white families. Similarly, any models \textit{deployed} on this data could disproportionally benefit white families by retrieving or summarizing more information for white families from the underlying data. We investigate the research question: are there systemic differences in ways notes about black and white families are written? We focus on two types of differences: quantitative (volume of data) and qualitative (word choice).

\begin{figure}
\centering
    \begin{tabular}{ccc}
    \hline
     & Num. & Avg. Word Count \\ 
     \hline
     \hline
    Black-assoc. Referrals & 50,490 & 1,485.13 \\
    White-assoc. Referrals & 65,393 & 1,323.96 \\
    Black-assoc. Cases & 7,525 & 50.81 \\
    White-assoc. Cases & 6,637 & 69.86 \\
    \end{tabular}
  \includegraphics[width=0.23\textwidth]{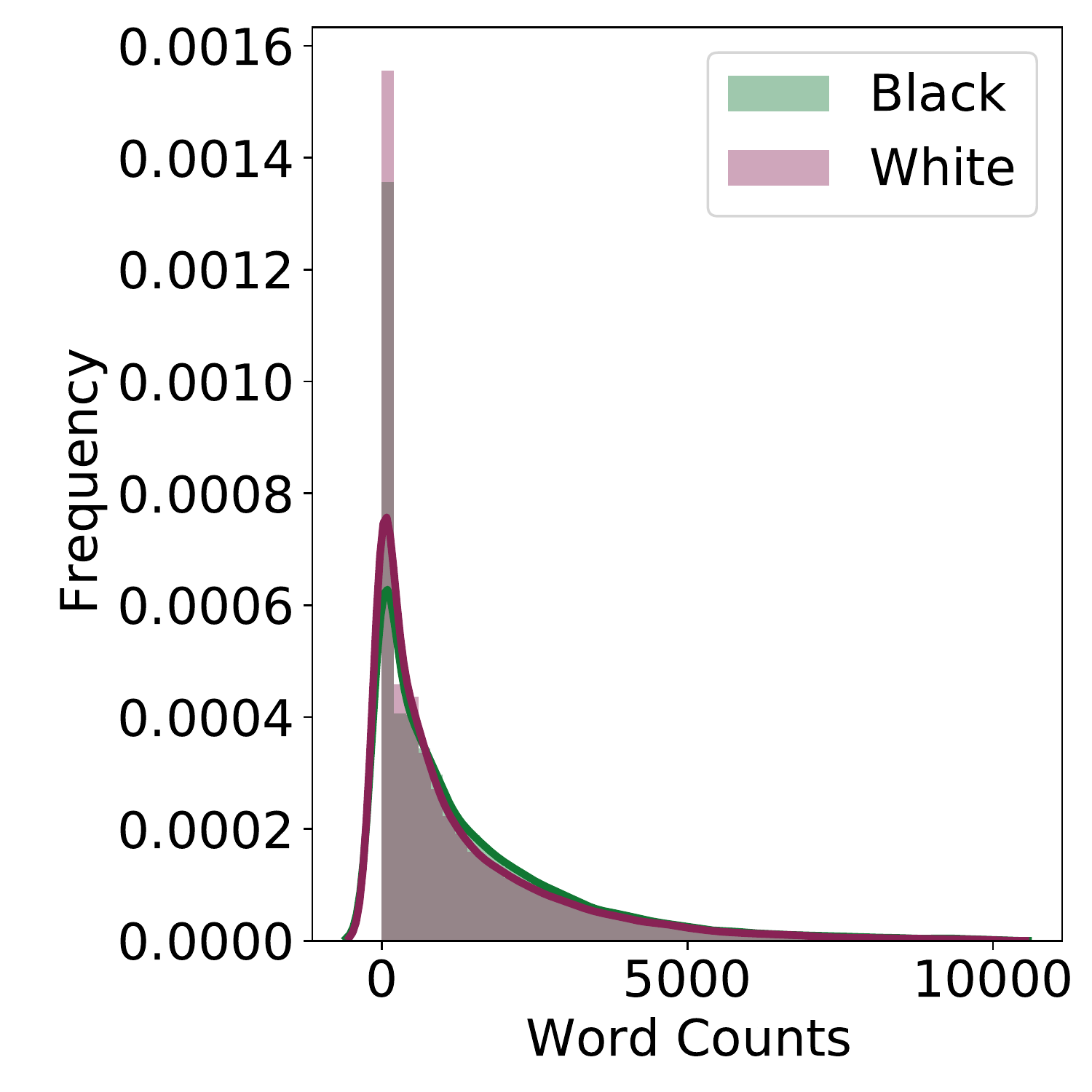}
\includegraphics[width=0.23\textwidth]{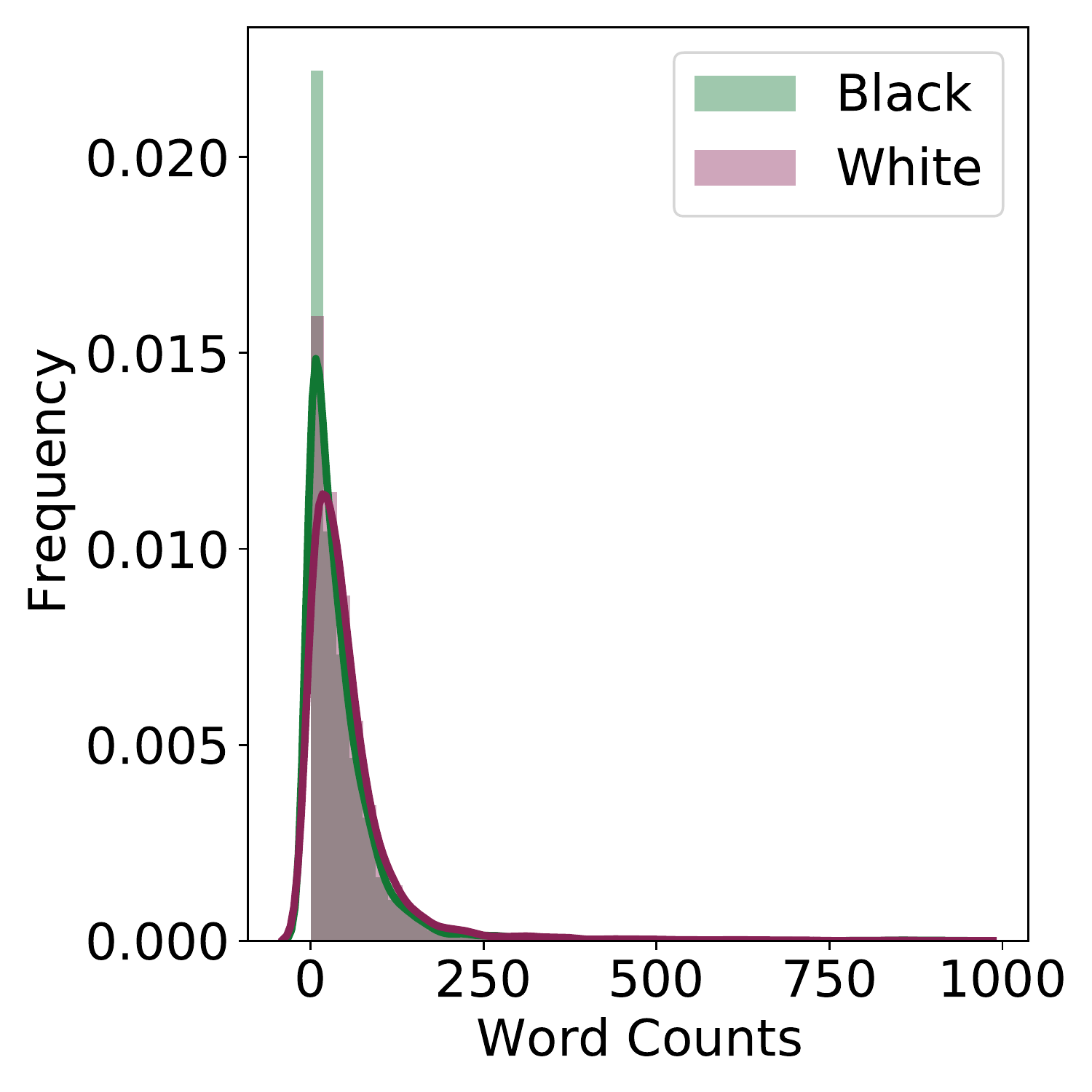}

\caption{Length of notes associated with cases and referrals by race. For cases, word counts are normalized by the number of days the case has been open. Left histogram shows word counts associated with referrals and right shows for cases (truncated for readability). There are not consistent length differences by race.}
  \Description{Two histograms, each with a series for black children and a series for white children. For both figures, the series have peaks at the same position and shapes are very similar.}
\label{fig:race_statistics}
\end{figure}

\subsection{Methodology}
\label{sec:data_statistics_methods}

\begin{table*}
    \centering
    \begin{tabular}{ccc|ccc}
    \hline
    Black-assoc. & Score & Nearest Match & White-assoc. & Score & Nearest match \\
    \hline
    \hline
    \textbf{Referrals} \\
she & 52.19 & he,that,stated & he & 54.64 & she,said \\
belt & 47.37 & spatula,paddle,spanked & heroin & 41.87 & Crack,ecstasy,crack-cocaine \\
her & 45.39 & his,him,and & PGF & 36.08 & PGGM,MGGM,GPGM \\
BM & 37.90 & BP,BPs,KCG & treatment & 36.16 & services,Outpatient,PND \\
bus & 30.95 & trolley,rides,stop & anxiety & 34.25 & unmedicated,schizophrenia,schizoaffective \\
shelter & 25.11 & motel,courthouse,AVAC & using & 27.45 & utilize,utilized,smoking \\
whooped & 23.96 & spanked,smacked,paddled & therapist & 26.05 & therapist-,school-based \\
    \textbf{Cases} \\
school & 56.80 & work,he,that & F & 130.67 & M,MGM,CW \\
housing & 42.01 & sub-iodized,HUD,housing/shelter  & parents & 59.26 & mother,mom,children \\
informed & 37.76 & stated,reported,also & drug & 37.65 & D+A,DOA,Screens \\
pass & 35.75 & re-issued,passed,fail & methadone & 36.55 & crack-cocaine,THC,Crack \\
\bottomrule
    \end{tabular}
    \caption{Words overrepresented in notes about black and white families computed using log-odds with a Dirichlet prior \cite{monroe_colaresi_quinn_2017}. Nearest-matching words that are not associated with the specified race are identified using cosine similarity of word embeddings. There are noticeable differences in topics and terminology in notes about children of different races. Commonly used DHS-specific acronyms are, BM: Birth Mother, BP(s): Birth Parent(s), KCG: Kinship Caregiver, PGF: Paternal Grandfather, PGGM: Paternal Great Grandmother, MGGM: Maternal Great Grandmother, D+A: Drugs and Alcohol. Others generally refer to service providers or common usage.}
    \label{tab:race_log_odds}
\end{table*}

To examine differences in data quantity, we compute the average number of words in contact notes for referrals and cases for black and white families.  We consider a case or referral to be black-associated if DHS metadata specifies >50\% of associated clients as ``Black or African American'' and analogously define white-associated. We do not include cases and referrals where neither of these criteria are  met (e.g. race of clients is unspecified, is other than black or white, or is mixed such that no one race is common to >50\% of clients). We exclude 4,075/18,237cases and 27,015/142,898 referrals using this criteria, where almost all of them have mixed or unknown race information, i.e. only 91/4,075  excluded cases have a client with a race other than Black, White, or Unknown specified.
Referrals have fixed time frames, but cases can span variable lengths of time. Thus, for cases, we normalize word counts by the number of days the case was marked as open in DHS systems. Normalizing by number of notes may seem more intuitive; however, we are interested in total volume of available data, not average note length, e.g., a case with many short notes may have lower average note length but more total data than a case with a few long notes.

To examine differences in data quality, we identify word-level language differences. We compute to what extent each word in the data is overrepresented in notes about black or white families as compared to all other notes using log-odds with a Dirichlet prior \citep{monroe_colaresi_quinn_2017}.
We then compare overrepresented words with common alternatives using 100-dimensional word embeddings trained over all 3.1M contact notes using skip-gram Word2Vec with a context window of 5. For each of the 100 most-overrepresented words, we identify the 3 words with the most similar (using cosine similarity) word embeddings that are not overrepresented (e.g. log-odds score $< 0$), discarding words that occur $<30$ times in the data set. 
These common alternatives offer perspective on what terminology note writers could have used, which can yield broader insights. For example, if hypothetically we found that the term ``deplorable'' was more common in black-associated notes, this result would be difficult to interpret. By comparing with common alternatives, we might find that ``disrepair'' is not black-associated but is used in similar contexts, which would suggest that note writers use more negative language when describing housing needs of black families.

\subsection{Results}
\Cref{fig:race_statistics} reports the average number of words and word-count histograms in contact notes associated with clients of different races on referrals and cases. Generally, there are not consistent differences by race in data set sizes. Although there tends to be more text data for white-associated cases (69.86/day compared to 50.81/day), there tends to be more text data for black-associated referrals (1,485.13 compared to 1,323.96).

\Cref{tab:race_log_odds} presents a subset of the 100 terms most overrepresented in case and referral notes about black and white families and their 3 nearest neighbors. 
Numerous words are overrepresented in both case and referral notes (duplicates not shown for brevity). Associations do reveal possible content and style differences along racial lines. Words common in notes about black families focus on behavior, punishment, and basic needs, while words common in notes about white families focus on drug use. Words related to women and female caregivers are also more common in notes for black families whereas notes for white families contain more references to male caregivers. While these metrics likely reflect differences in events and reasons that families are referred to CPS, there are also racial associations in some near-synonyms, likely more reflective of style than content: notes about black children use ``whooped'' over ``spanked'', and ``informed'' over ``stated''.
Manual examination of notes containing overrepresented terms shows that note writers often directly and indirectly quote clients and sources they interview, for instance notes about black families often contain African American English (AAE).
Some notes quotation marks, e.g., \textit{C reports that F tells him ``sit down and listen''}, but language likely still reflects terms used by clients and sources even when not marked with quotation marks, e.g., \textit{he said he was trying to get out of the room so he can go outside}.\footnote{Examples were modified to preserve privacy, not directly drawn from the data.}
Thus, terminology differences result both from terms used by note writers themselves and by people they quote. 
It is well documented that such subtle differences in language can lead to harmful biases in downstream tasks, e.g., off-the-shelf NLP models for toxic language classification are more likely to falsely classify AAE as offensive
\citep{davidson-etal-2019-racial,sap-etal-2019-risk,field2021Survey}.
While much scope remains for further investigation into the origins and effects of these differences as well as closer examination of language variance beyond word-level metrics, these word statistics do suggest that there are systemic differences in notes about children of different races which could be absorbed and amplified by NLP models. We explore some such models in the following sections.

\section{Racial Disparities in Risk Assessment}
\label{sec:risk_assessment}

Public agencies are increasingly turning to algorithmic models with the goal of improving the consistency and accuracy of time-sensitive high-stakes decisions \citep{studentsatrisk, chouldechova2017fair,vaithianathan2017developing,Raghavan2020Mitigating,Saxena2020Human,dhs_2019}. These models reflect evolution from operator-driven checklists derived from regressions to machine-learning methods that draw on hundreds of pieces of information \citep{mcnellan2022evidence}. Risk assessments tools in general have been criticized for failing to account for relevant individual context and for automating biases in the data \cite{whittaker2018ai, eubanks2018automating, roberts2019digitizing}. In a survey of CPS algorithms, \citet{Saxena2020Human} find that current tools rely on structured tabular data and suggest that augmenting the data features with natural language could be one way to incorporate context, improve model performance, and reduce bias.
Though they dispute this suggestion in follow up work \citep{moon2023rethinking}, the initial suggestion evidences how contact notes may appear useful data for risk prediction: they generally contain numerous details not captured in structured data.
However, their incorporation could exacerbate many of the concerns around risk assessment: text is written by people and reflects their perceptions of events, which may or may not accurately reflect reality \cite{eberhardt2020biased,moon2023rethinking}.
There are numerous risks associated with incorporating text notes into risk predictive models, including reducing transparency, overfitting, increasing surveillance and privacy violations. In this section, we focus on algorithmic bias as one specific risk, and we provide additional discussion in Section~\ref{sec:discussion}.

In the anonymous county, DHS uses a predictive risk assessment tool to aid in call-screening. For an incoming referral, the tool presents call-screening staff with a score from 1 to 20 that aims to reflect the likelihood that the child will be placed (removed from home) within 2 years conditional on the referral being screened in. While the model has undergone changes, the original version was a logistic LASSO that selected 71 features from  $>800$ variables providing demographics, past welfare interaction, public welfare, county prison, juvenile probation, and behavioral health information on all persons associated with each referral.
Some features are derived from previous interactions with the child welfare system, (e.g. the number of previous referrals associated with people on the new referral).

\citet{chouldechova18a} show that call-screening tools can exhibit miscalibration by race, and \citet{Cheng2022} show that without caseworker oversight, they can result in a much higher screen-in rate for black children than white children. In this section, we integrate text features into the county's existing tool and analyze the impact on model performance, focusing on racial disparities. We examine several related research questions: Does integrating text features:

\begin{enumerate}[noitemsep,topsep=0pt,parsep=0pt,partopsep=0pt]
\item increase model performance disparities for black and white children?
\item increase model miscalibration with respect to race?
\item increase the proportion of black children flagged as high risk by the model?
\end{enumerate}

\subsection{Methodology}
\paragraph{Data and Features}
We base our models on the same data and features used in the original version of the model, which encompasses referrals screened in for investigation from April 2010 to July 2014. 
The basic data unit is a referral–child pair: if a maltreatment referral had multiple children or if the same child was included in multiple referrals, we treat each unit as a separate data point.
We use pre-constructed test and training splits, which were constructed to ensure no overlap in children or referrals between the training and test set, and we reserve 10\% of the training set as a validation set.
Although all child–referral observations contain structured features based on the current referral, not all families have had prior interactions with DHS and prior interactions could have been expunged. Thus, not all observations contain associated text notes. 
Our final data set consists of 28,769 training instances, 7,893 of which contain text data, and 14,417 test data points, 4,133 of which contain text data.

Models are trained to predict out-of-home placement within 2 years using structured and text features. As structured features, we use the same 818 features as early versions of the model. For text features, for each child in each referral, we pull contact notes for prior cases and referrals where the child was listed as a client, restricted to notes from the previous 365 days that contain the child's first name. We restrict data to the preceding year based on suggestions from DHS employees and early experiments with the validation set, which suggested that data from this time frame most improved model performance when compared to longer or shorter time frames. We preprocess the text by expanding common acronyms (e.g., F = father) using a list manually curated in consultation with DHS. We further remove first and last names of clients as listed in associated metadata and mask any additional people and location named entities identified using SpaCy.

\paragraph{Models}
We examined four standard classification models: logistic regression, random forest, GatedCNN (a convoluted neural network (CNN)-based neural classifier shown to perform well on a similar task over medical text) \citep{ji-etal-2021-medical}, and a RoBERTa-based neural classifier \citep{Liu2019RoBERTaAR}. As the random forest model achieved the best classification performance, likely due to data imbalance and limited high-dimensional training data, and it is most similar to previously investigated call-screening tools \citep{chouldechova18a}, we describe implementation and results from this model here and describe other models in Appendix~\ref{sec:appendix_nlp_models}.
We report results for the random forest model without text features (structured) and with text and structured features (hybrid). For the structured model, the feature-input is 818 dimensional structured feature vectors. To incorporate text features, we first trained a text model using a logistic regression classifier with TF-IDF-weighted bag-of-words features and a 10,000 word vocabulary.
We then took the 500 words with the highest learned coefficients and the 500 words with the lowest (most negative) coefficients and constructed TF-IDF features from this refined 1,000 word vocabulary. We then concatenated these features with the 818 structured features, constructing 1,818-dimensional feature vectors.
We do this feature selection because we found in early experiments over the validation set that concatenating all text features with the 818 structured features caused the model to ignore the structured features.
All random forest classifiers used 500 trees.

\begin{table*}
    \centering
    \begin{tabular}{c|c|c|c|c|c|c|}
    Test set &    Model & AUC & Avg. Pos Score & Avg. Neg Score & FPR & FNR \\
        \hline
    \multirow{2}{*}{Full} & Structured &  $ 75.77 \pm 0.02 $ & $ 13.85 \pm 0.00 $ & $ 8.70 \pm 0.00 $ & $ 19.59 \pm 0.01 $ & $ 43.96 \pm 0.05 $ \\
& Hybrid &  $ 76.27 \pm 0.02 $ & $ 13.94 \pm 0.00 $ & $ 8.69 \pm 0.00 $ & $ 19.53 \pm 0.01 $ & $ 43.60 \pm 0.07 $ \\
        \hline
     \multirow{2}{*}{Examples with notes} & Structured &  $ 69.83 \pm 0.03 $ & $ 14.54 \pm 0.01 $ & $ 11.19 \pm 0.01 $ & $ 31.78 \pm 0.06 $ & $ 40.51 \pm 0.09 $ \\
& Hybrid &  $ 71.88 \pm 0.04 $ & $ 15.84 \pm 0.01 $ & $ 13.24 \pm 0.01 $ & $ 40.34 \pm 0.08 $ & $ 29.80 \pm 0.13 $ \\
\bottomrule
    \end{tabular}
    \caption{Performance of structured and hybrid models trained and evaluated on predicting out-of-home placement. The feature inputs to the structured model are the tabular structured data, and the feature inputs to the hybrid is both the structured data and contact notes. Avg. Pos/Neg Score report the average predicted risk scores for true positive (placement occurred) and true negative (no placement) test data, where risk scores are computed by bucketing test predictions into ventiles. \emph{Top:} Differences in model performance across the full test set ($n =14,417$) are small. \emph{Bottom:} Differences across the test set that contains text data ($n= 4,133$) show reductions in false negatives, but not in false positives.}
    \label{tab:performance_supervised}
\end{table*}

\begin{table*}
    \centering
    \begin{tabular}{c|c|c|c|c|c|c|}
       Test Set & Model & AUC$_{black -white}$ & Avg. Pos Score$_{b-w}$ & Avg. Neg Score$_{b-w}$ & FPR$_{b-w}$ & FNR$_{b-w}$ \\
        \hline
   \multirow{2}{*}{Full} & Structured &$ -0.26 \pm 0.04 $ & $ 0.35 \pm 0.01 $ & $ 0.66 \pm 0.01 $ & $ 1.09 \pm 0.06 $ & $ -2.87 \pm 0.13 $ \\& Hybrid &$ 0.72 \pm 0.04 $ & $ 0.34 \pm 0.01 $ & $ 0.31 \pm 0.01 $ & $ 0.35 \pm 0.06 $ & $ -4.00 \pm 0.16 $ \\
        \hline
   \multirow{2}{*}{w/ notes} & Structured &$ -1.83 \pm 0.06 $ & $ 0.14 \pm 0.01 $ & $ 0.72 \pm 0.01 $ & $ 2.72 \pm 0.10 $ & $ -0.92 \pm 0.18 $ \\
& Hybrid &$ -1.05 \pm 0.08 $ & $ 0.35 \pm 0.01 $ & $ 0.68 \pm 0.01 $ & $ 6.17 \pm 0.13 $ & $ -4.44 \pm 0.22 $ \\
\bottomrule
    \end{tabular}
    \caption{Performance disparities by race for the structured and hybrid models. The predictive disparities are largely comparable for the structured and hybrid models. We do not therefore find evidence that incorporation of text features increases aggregate measures of algorithmic unfairness in this setting. Raw performance values are reported in Appendix~\ref{sec:additional_model_metrics}.
    }
    \label{tab:performance_supervised_by_race}
\end{table*}

\paragraph{Fairness Metrics}
In order to examine (1), if text features increase model performance disparities, we examine the difference in model predictions with and without text features for black and white children (e.g., accuracy equity and error rates) using several performance evaluation metrics: area under the ROC curve (AUC), false positive rate (FPR), false negative rate (FNR), and the raw model scores outputted for children who were placed out of home (Average Positive Score) and who were not (Average Negative Score).
In order to both provide realistic estimates of how these systems may operate when deployed in practice and highlight differences in performance when incorporating text features, we report metrics over the full test set and over only the subset of the test data that contains text features. When computing metrics requiring a classification decision (e.g.,~FPR, FNR), we consider the 25\% of test data with the highest raw output scores as having positive predictions. This percentage corresponds to the mandatory screen-in threshold that has been used by DHS.
Additionally, we conduct bootstrap sampling with the training data and report average metrics and standard error values computed from 100 samples.

To evaluate (2), if text features increase model miscalibration with respect to race, we plot true placement rate for each bracket of model-predicted risk score, following \citet{chouldechova18a}.
Both (1) and (2) compute results based on out-of-home placement values, but this proxy outcome may itself reflect racial bias. Thus, we examine (3), if text features increase the proportion of black children flagged as high risk by the model, by comparing the averages and distributions of predicted risk scores for black and white children.
Overall, these metrics aim to capture if incorporating text features into the tool is likely to disproportionally harm black families referred to CPS by increasing the chance their referral is investigated or by creating racial disparities in model errors.

\subsection{Results}

\Cref{tab:performance_supervised} reports overall performance of the structured and hybrid models. Changes in AUC and in predicted risks scores are small, though the hybrid model does consistently outperform the structured model.
The improvements are largely driven by increases in the risk scores for data points where out-of-home placement occurred, which decrease the false negative rate.
However, the incorporation of text features does not decrease the false positive rate for data points with text features, nor result in lower risk scores for data points where out-of-home placements did not occur.
In general, the model interprets the presence of associated text as an indicator of risk, as test data points with text are assigned higher scores by the hybrid model than the structured model, regardless of whether or not out-of-home placement occurred.

\begin{figure}
    \centering
        \includegraphics[width=0.35\textwidth]{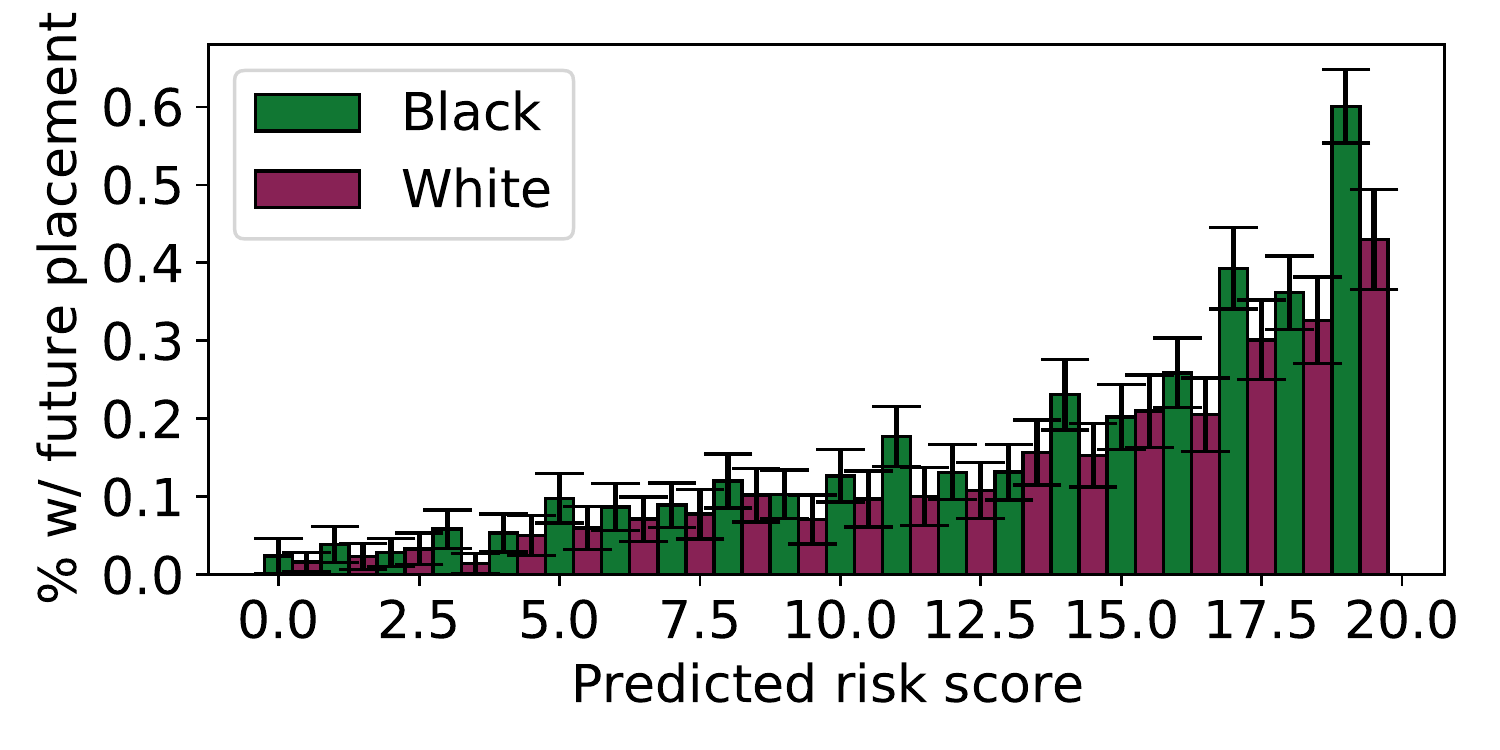}
    \includegraphics[width=0.35\textwidth]{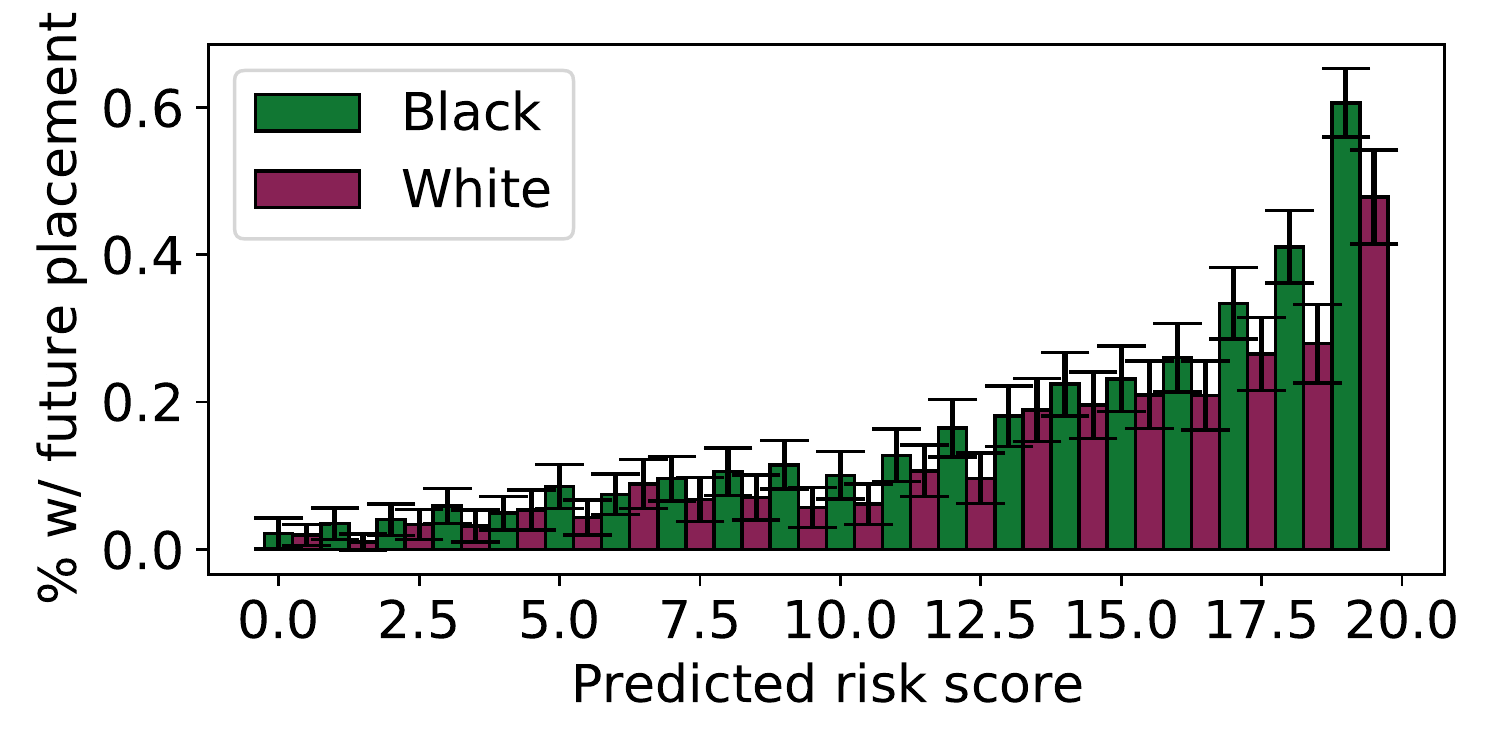}
    \caption{Calibration plots using structured (top) and hybrid (bottom) features  (\citet{chouldechova18a} provides details on computing/interpreting calibration plots). We infer predicted risk scores by grouping the full data set into ventiles, but only display data points for black and white children in these figures. Both the structured and hybrid models display signs of miscalibration in the highest ventile.}
    \Description{Two bar charts. Each has one series for black children and one series for white children. Each bar value shows the percentage of children actually placed out of home out of all the children assigned a particular risk score by the model. For the highest risk score, the right-most value of the series, both charts show that a higher percentage of black children were placed out of home than white children.}
    \label{fig:callibration}
\end{figure}

 \Cref{tab:performance_supervised_by_race} reports results for question (1), if text features increase performance disparities. Differences in model performance for black and white children are small overall, and the hybrid model does not show greater performance disparities than the structured model. In identifying referrals without future out-of-home placement (e.g. FPR), the hybrid model slightly reduces disparities (1.09 to 0.35). In identifying referrals with out-of-home placements (e.g. FNR), the hybrid model slightly improves performance for black children more than white children.

\Cref{fig:callibration} addresses question (2) and compares risk scores using calibration plots. Both the structured and hybrid models display signs of miscalibration in the highest risk bracket. Specifically, a higher percentage of black children assigned the highest risk were placed out-of-home than white children, but the hybrid model does not appear any more miscalibrated than the structured model.

\begin{figure}
    \begin{tabular}{ccc}
    \hline
     Model & $\%race_{b-w}$ & $\%flagged_{b-w}$ \\ 
     \hline
     \hline
     Structured & $ 3.18 \pm 0.06 $ & $ 13.30 \pm 0.11 $ \\
     Hybrid & $ 2.73 \pm 0.07 $ & $ 12.50 \pm 0.12 $ \\
    \end{tabular}
  \includegraphics[width=0.23\textwidth]{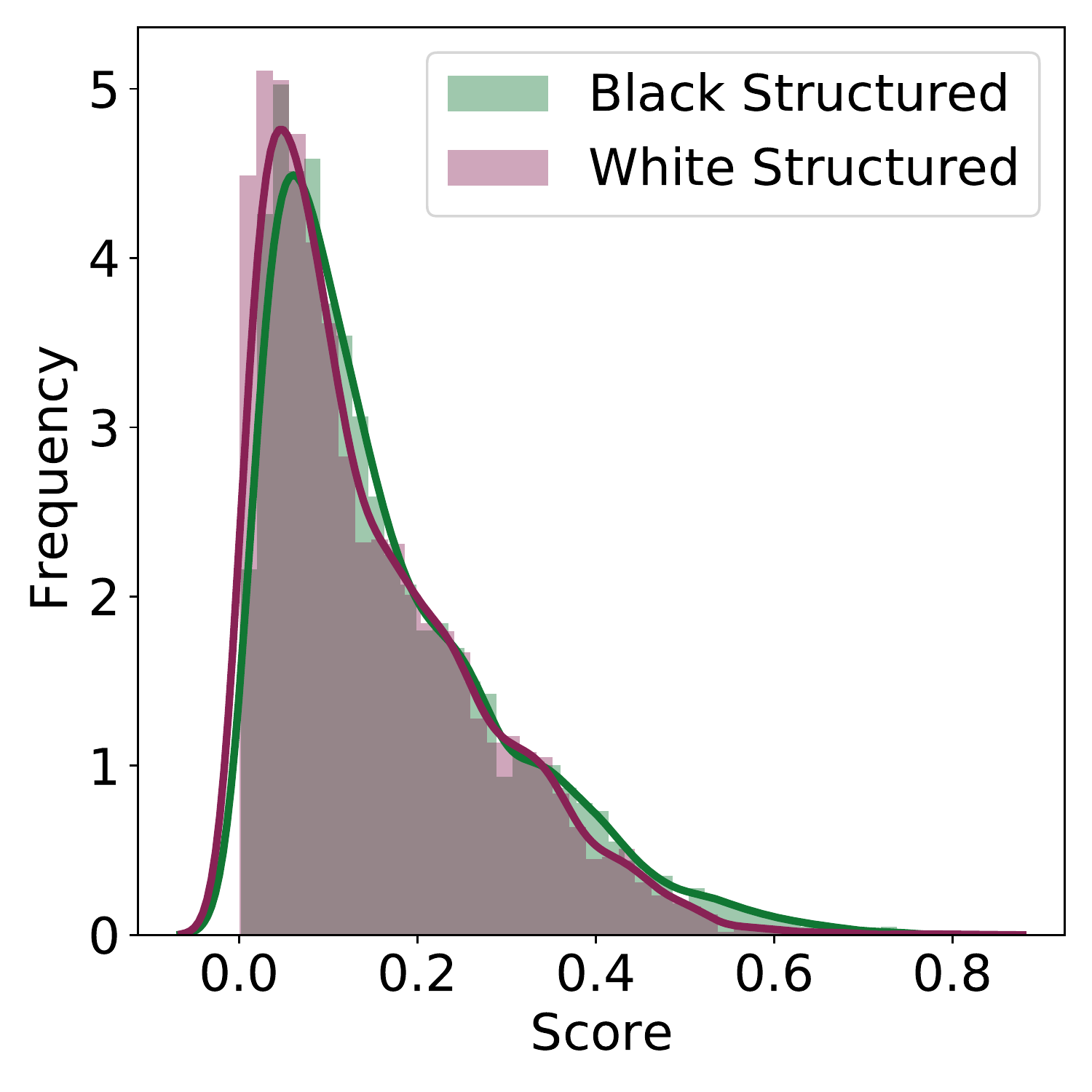}
  \includegraphics[width=0.23\textwidth]{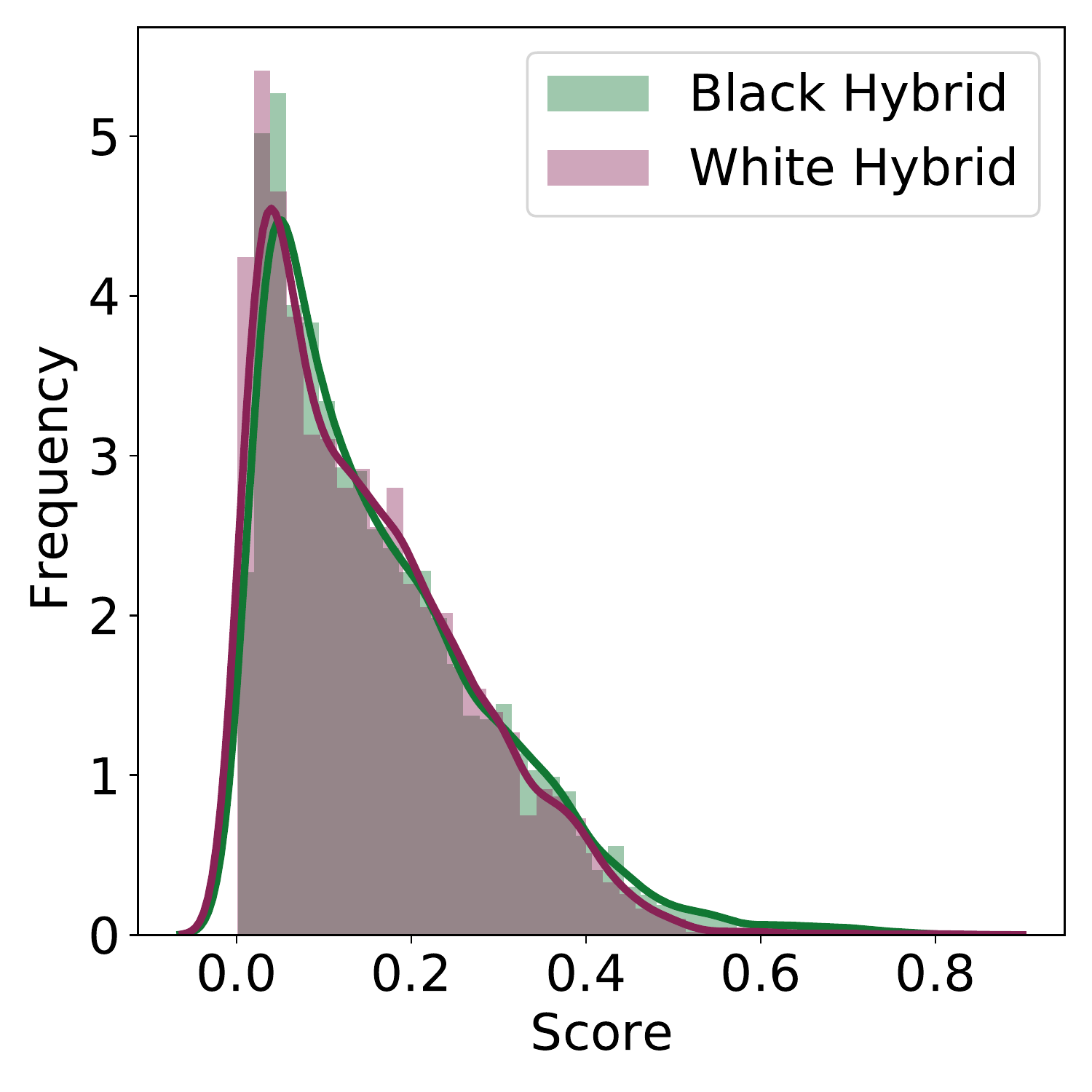}

\caption{\emph{Top:} difference in the percent of children of each race (black - white) that are flagged as high-risk under the specified model (\%race), difference in the racial composition (black - white) of those who are flagged as high risk (\%flagged). The hybrid model shows a reduced disparity in the percentage of black children who are flagged as high-risk. \emph{Bottom:} histograms of the raw scores outputted by each model, divided by race. Under the structured model (left), score predictions for black children are right-shifted compared to white children. Under the hybrid model (right), the score distributions are nearly identical. }
\Description{Two histograms, each with a line for black children and a line for white children. In the left histogram, the peak for white children is slightly to the left of the peak for black children. In the right histogram, the peaks are nearly identical.}
\label{fig:race_raw_scores}
\end{figure}

Finally, to address question (3), \Cref{fig:race_raw_scores} displays the percent of children of each race that are flagged as high-risk  as well as the racial composition of those that are flagged for both models, allowing us to examine if the hybrid model is liable to increase the number of black families involved in CPS. Under both models, a higher percentage of black children are flagged as high-risk relative to white children, but the hybrid model reduces this disparity. This reduction is also visible in the histograms of raw model scores: under the structured model, scores for black children are right-shifted compared to white children, while under the hybrid model, the score distributions are nearly identical.

Overall, the changes in model performance when text features are incorporated suggest that text features do not increase aggregate measures of racial disparities in model predictions and may actually improve them.
Prior work has shown racial bias is undoubtedly a concern in deploying algorithms in CPS settings \citep{chouldechova18a,Cheng2022}. 
However, in the context of NLP research priorities, our results suggest that model debiasing may not be useful in this particular task, and focus on algorithmic bias may be over-estimating perceived risks and distracting from true risks, like reducing transparency and risking privacy violations.
Our work offers corroborating empirical evidence to more theoretical discussions on how NLP research on debiasing can be misplaced \citep{blodgett-etal-2020-language}.
Nevertheless, although we do not find evidence of increased bias in this specific experimental setup, we emphasize that this finding should not be interpreted as a generalizable lack of racial bias in the contact notes or in the predictive models used on those notes; results may differ under different models or subsets of the data.

\section{Racial Disparities in Information Extraction Systems}
\label{sec:ner}

While risk assessment models reflect technology already in use in CPS settings, for text processing, CPS agencies are more actively interested in developing NLP systems that focus on aiding CPS workers in retrieving and organizing information rather than direct decision-making.
For example, in 2018 one agency solicited proposals for tools that mined information from text, including identifying family support systems (e.g., names of extended family members who can provide care) and issues of concern (e.g., substance use, intimate partner violence) \cite{dhsRFP2018}.
In the research community, \citet{perron2019detecting} provide an example: using NLP to determine if a case note mentions a substance-related problem; in follow up work, \citet{perron2022text} similarly explore using a  rule-based entity recognition system for identifying opioid mentions.

\begin{table*}[ht]
    \centering
    \begin{tabular}{ccccccccccccc}

  & & \multicolumn{5}{c}{Full Names} & & \multicolumn{5}{c}{First Names} \\
    \cline{3-7} \cline{9-13}
    
     &  & \small{\# names-} &  & & \small{FlairNLP} & \small{FlairNLP} &  & \# \small{names-} & &  &  \small{FlairNLP} & \small{FlairNLP}  \\
    & & \small{notes} & \small{SpaCy} & \small{NLTK} & \small{(ConLL)} &  \small{(OntoNotes)}  & & \small{notes} & \small{SpaCy} & \small{NLT}K & \small{(ConLL)} &  \small{(OntoNotes)} \\
      \hline
      \hline
\multirow{3}{*}{\rotatebox{90}{\small{Referrals}}}
& Black            & 95K  & 78.3\%  & 83.5\%  & 98.0\%  & 95.6\%  & & 314K & 68.0\%  & 83.8\%  & 97.2\% & 96.0\% \\
& White            & 108K & 83.4\%  & 86.9\%  & 99.1\%  & 97.2\%  & & 368K & 76.5\%  & 88.4\%  & 98.3\% & 97.3\% \\
& B - W &      & -5.1\%  & -3.4\%  & -1.1\%  & -1.6\%  & &      & -8.5\%  & -4.6\%  & -1.1\% & -1.3\% \\

\hline
\multirow{3}{*}{\rotatebox{90}{\small{Cases}}}
& Black            & 858K & 72.85\%  & 78.61\%  & 97.18\%  & 94.67\%  & & 6.7M & 61.47\%  & 81.67\% & 96.24\% & 95.21\% \\
& White            & 538K & 77.99\%  & 83.16\%  & 98.87\%  & 96.76\%  & & 4.2M & 72.79\%  & 86.68\% & 97.99\% & 97.06\% \\
& B - W &      & -5.14\%  & -4.55\%  & -1.69\%  & -2.09\%  & &      & -11.32\% & -5.01\% & -1.75\% & -1.85\% \\
\hline
    \end{tabular}
    \caption{NER model recall at recovering names in structured data from contact notes. Across all models, recall is higher for names of white clients than black clients. All performance differences between names of white and black clients are statistically significant.}
    \label{tab:ner}
\end{table*}

We investigate racial disparities in named entity recognition (NER) and coreference resolution, which are necessary components of information extraction systems. While a domain-specific model might identify substance use, NER and coreference are necessary for resolving who in a case is involved: substance abuse for a parent is a different situation than for a child.
There are also specific imminent use cases: although the county maintains structured records of people involved in cases, they are not always up-to-date. In the event of a crisis situation and a court orders a child to be removed from their home, NER models could aid a caseworker in finding immediate kinship placement options over a non-kinship or group home.
For coreference, we focus on use cases previously explored for processing expert-written notes in the medical domain. This framework aims to directly identify mentions of \textit{People}, \textit{Problems}, \textit{Treatments}, and \textit{Tests} in addition to coreference links between them \citep{uzuner20112010}.
Our focus on these two tasks is additionally motivated by conversations with DHS employees about what NLP tools they would find useful and are actively considering deploying.

Furthermore, NER and coreference resolution are both well-established NLP tasks with existing off-the-shelf-deployable models, which makes their potential deployment a realistic scenario.
While prior research has investigated gender or racial bias in these tasks over synthetic data \citep{rudinger-etal-2018-gender,zhao-etal-2018-gender,mishra2020assessing}, to the best of our knowledge, no work has explored more realistic settings.
Here, we investigate: do existing NER and coreference resolution models exhibit performance disparities over notes about black and white families?
This question aims to capture possible direct harms that could occur from model deployment. An NER system that has higher recall for names of white clients than black clients could result in black children being placed initially in non-kinship homes more often, even when kinship placement options were available. In contrast, a coreference system that has higher recall at identifying \textit{Problems} for black families and \textit{Treatments} for white families could unfairly highlight risks for black families and supports for white families.

\subsection{Methodology}
\label{sec:ner_methods}

\paragraph{NER}
We evaluate four NER models: SpaCy ``en\_core\_web\_sm'' default NER model (a transition-based parser trained on OntoNotes), NLTK's currently recommended NER module (a MaxEnt classifier trained on the ACE corpora) and two neural models based on document-level XLM-R embeddings, one trained on CoNLL-03 and one trained on OntoNotes implemented with the fairNLP packages \citep{schweter2020flert}. We selected these models to cover a range of training data sets and architectures and also due to their ease of implementation, which makes them more likely for CPS agencies to adopt. For each model, we examine entities tagged with \textit{PER} or \textit{PERSON}. 

For evaluation, we use existing county records of clients involved in cases and referrals as gold named entities, and we evaluate the ability of NER models to recover these names from contact notes. We conduct this analysis at a case/referral-client-note level, which best replicates a realistic scenario: e.g. a CPS worker may search through notes on a case to find all information about possible caregivers.
We again focus on black and white clients and do not include clients who are of mixed race or other races.
Our primary evaluation metric is recall, as output from an NER system would be read by a CPS worker who could easily disregard any incorrect names.
CPS workers often use acronyms, so many clients listed on cases are not mentioned at all by name in contact notes. To account for this, we searched for the client's name in the relevant notes and only computed model recall over names that appeared in the text.
In practice, NER models are more useful for identifying people not already listed in structured data; however, names in structured data offer us an existing evaluation set that is reflective of how well NER models can capture people described in notes.
We focus on comparing model recall for names of black clients with names of white clients.

\paragraph{Coreference}
We examine two neural coreference systems. The first is an extremely common coreference architecture where an encoder, mention detector, and antecedent linker are trained end-to-end \citep{lee-etal-2018-higher}. The second is a state of the art variant of \citet{lee-etal-2018-higher} that relies only on the start and endpoint of the span \citep{Kirstain2021CoreferenceRW}. For both models we use a SpanBERT encoder \citep{joshi2020spanbert} and train on OntoNotes \citep{hovy2006ontonotes}. As OntoNotes differs greatly from CPS contact notes, we employ continued training on more similar domains, which is a standard approach to overcoming established domain-transfer challenges for coreference \citep{xia-van-durme-2021-moving}. We focus on two settings: (1) continued training on a limited subset of annotated contact notes (keeping a test set held-out) and (2) continued training on clinical notes from the i2b2/VA corpus \citep{uzuner20112010}, which also consists of expert-written notes and uses the same annotation scheme as the annotated contact notes.

As evaluation and continued training data, we use a data set of 200 contact notes annotated for coreference resolution in prior work \citep{gandhi2022mention}, with train/dev/test sets of sizes 100/10/90.
These notes were annotated using the same schema as prior work on medical notes \citep{uzuner20112010} and examples include, \textit{People}: names, pronouns, acronyms; \textit{Problems}: housing insecurity, physical abuse, substance use; \textit{Tests}: structured assessments carried out by CPS workers, records of school attendance; \textit{Treatments}: housing services, sources of support, treatment plans. 
For all models, we report evaluation over the held-out test set over notes where the majority of clients are white (31 notes) or black (49 notes), using the same definitions as in Section~\ref{sec:data_statistics_methods}.
For models without continued training where we do not need to reserve data for training, we additionally report results over all black-majority (87) and white-majority (78) notes in the annotated data.
We adopt the standard approach of averaging three coreference metrics: $\text{MUC}, \text{B}^3, \text{CEAF}_{\phi_4}$.


\subsection{Results}

\Cref{tab:ner} presents recall scores for all NER models. Recall scores for flairNLP models in particular are high. For all models, recall is significantly higher for names of white clients than for names of black clients.
These findings are similar to the analysis of synthetic data conducted by \citet{mishra2020assessing}, who demonstrate that NER performance for several models is higher over white first names than black first names. Our work shows that this finding holds in a real and much a larger data set and over full names as well as first names.
In examining names missed by the models, models typically fail to identify uncommon spellings, such as ``Emilie'' or names that are also non-proper nouns, such as ``Precious'' or ``Ruby''.\footnote{To preserve privacy, these examples are fabricated based on general patterns observed in the data, not real names of clients.}
Although performance differences are smaller for better-performing models, we examine a very large data set, so even a 1\% score difference reflects 1000s of name-instances and is statistically significant. 

In contrast, the test set of coreference annotations is quite small. Collecting more data is extremely difficult, as annotations are time-consuming and require domain expertise \citep{gandhi2022mention}. While we discuss performance in the absence of a larger data set, we caution that results are generally not statistically significant.
From \Cref{tab:coref}, for the two off-the-shelf-style models that are not trained on in-domain data, \citet{Kirstain2021CoreferenceRW} and \citet{lee-etal-2018-higher} (i2b2/VA), performance is poor, and manual examination of outputs suggests the models are not accurate enough to be usable. Training on in-domain data improves model performance (\citet{lee-etal-2018-higher} (contact notes)), but reverses the direction of model bias: while \citet{Kirstain2021CoreferenceRW} and \citet{lee-etal-2018-higher} (i2b2/VA) perform slightly better or equal on notes about black-majority families, \citet{lee-etal-2018-higher} (contact notes) performs better on notes about white-majority families.

\begin{table}
    \centering
    \begin{tabular}{cccc}
    \hline
     &  Kirstain & \citet{lee-etal-2018-higher} & \citet{lee-etal-2018-higher} \\
     &   et al. \cite{Kirstain2021CoreferenceRW} & (i2b2/VA) &  (contact notes) \\
      \hline
      \hline
       Black & 58.82\% & 43.96\% & - \\
       White & 57.11\% & 41.92\% & -  \\
       B - W & 1.71\%  & 2.04\% & - \\
\hline
Black    & 56.98\%  &  43.58\%	& 66.81\% \\
White    &  57.12\%	 &	41.22\% & 68.24\%  \\
B - W    &   -0.14\% &  2.36\%  & -1.43\% \\
\bottomrule
    \end{tabular}
    \caption{Average F1 for coreference models in annotated set of 165 contact notes (top) and in held-out test set of 80 notes (bottom). (i2b2/VA) and (contact notes) indicate the data set used for continued training. There is no statistically significant difference in performance over contact notes about  majority-white and majority-black families for any models.}
    \label{tab:coref}
\end{table}

\begin{table*}
    \centering
    \begin{tabular}{ccccccccccccc}
    \hline
    & \multicolumn{4}{c}{\# of Entities} & \multicolumn{4}{c}{\citet{lee-etal-2018-higher} (i2b2/VA)} & \multicolumn{4}{c}{\citet{lee-etal-2018-higher} (contact notes)} \\\cline{2-5} \cline{6-9} \cline{10-13}
     &  Per. & Treat. & Test & Prob. & Per. & Treat. & Test & Prob. & Per. & Treat. & Test & Prob.\\
      \hline
      \hline
     Black   & 424 & 269 & 195 & 290 & 50.81\% & 54.59\% & 59.39\% & 54.56\% & - & - & - & - \\
     White   & 366 & 238 & 163 & 248 & 48.04\% & 54.57\% & 59.43\% & 51.55\% & - & - & - & - \\
     B - W   &     &     &     &     & 2.77\% & 0.02\% & -0.04\% &  3.01\% & - & - & - & - \\
\hline
      Black   & 59 & 42 & 30 & 40 & 51.14\% & 53.17\% & 55.61\% & 54.21\% & 82.21\% & 83.89\% & 90.22\% & 83.59\% \\
      White   & 39 & 29 & 24 & 29 & 46.11\% & 52.81\% & 54.75\% & 50.17\% & 81.32\% & 86.08\% & 89.19\% & 81.64\%\\
      B - W    &   &    &    &    & 5.03\%  & 0.36\%  & 0.86\%  & 4.04\%  & 0.89\%  & -2.19\% & 1.03\% & 1.95\%\\
\bottomrule
    \end{tabular}
    \caption{Average recall for coreference models for entity types over 165-note annotated data (top) and  held-out 80-note test set (bottom). For both models, recall of \textit{Problems} is higher for notes about black-majority families than white-majority ones.
    For \citet{lee-etal-2018-higher} (contact notes), recall of \textit{Treatments} is higher for white majority families.}
    \label{tab:coref-concept-small-test-set}
\end{table*}

\Cref{tab:coref-concept-small-test-set} provides a finer-grained breakdown of model performances, presenting model recall of each entity type for the \citet{lee-etal-2018-higher} (i2b2/VA) and \citet{lee-etal-2018-higher} (contact notes) models, which are trained on data annotated with these entities.
Both configurations exhibit stronger recall over \textit{Person} and \textit{Problem} entities for notes about black-majority families. \citet{lee-etal-2018-higher} (contact notes) additionally achieves better recall of \textit{Treatment} entities for notes about white-majority families.
The origins of these disparities is difficult to untangle: they could reflect correlations between the types of \textit{Problem}/\textit{Treatment} entities mentioned in notes about families of different races and the types of entities that models recall better, e.g. from Section~\ref{sec:data_statistics}, notes about white families tend to discuss substances abuse more than basic needs. They also could reflect biases absorbed from external model training data or biases from ways notes are written. 
Nevertheless, regardless of origin, a coreference system that retrieves a smaller proportion of \textit{Problem} entities in white-majority contact notes and a smaller proportion of \textit{Treatment} entities for black-majority notes might lead to disproportionate focus on risks for black families and supports for white families.

\section{Discussion}
\label{sec:discussion}

Despite numerous associated risks, in reality CPS agencies are actively seeking to deploy NLP tools. At the same time, NLP benchmark data sets are not sufficient for assessing model performance.
While much literature has critiqued risk assessment systems \citep{eubanks2018automating,moon2023rethinking}, little work has explored the impact of NLP in these tools, nor potential biases in assumedly more benign information extraction systems.
We find significant racial bias in NER systems and possible biases in coreference systems (Section~\ref{sec:ner}), but we do not identify racial bias as a core concern with integrating text features into an existing risk assessment system (Section~\ref{sec:risk_assessment}). We additionally document evidence of different language use by race in contact notes (Section~\ref{sec:data_statistics}), which could have impacts on any models trained or deployed over this data.

Our results suggest that risks other than racial bias are more important considerations regarding incorporating text into risk assessment systems.
Through an in-depth analysis of contact notes, \citet{moon2023rethinking} discourage using this data in risk assessment models and emphasize ways CPS systems deprive families of agency, over-surveil parents, and problematize the data collection processes. Our finding that the models we train tend to assign higher risk scores to any referrals with pre-exisiting text data (Section~\ref{sec:risk_assessment}) also suggest that incorporating text data could increase over-surveillance by encouraging repeated investigation of families that have already been involved in CPS.
There are also risks that come from text as a data type. High-dimensional text data are liable to increase overfitting to ``proxy'' values used to train risk models. Also, text data are extremely difficult to anonymize. Because algorithmic systems are typically built by external researchers or contractors, text processing requires sharing it externally, which reduces the privacy of families and increases the potential harms associated with any data breaches. Finally, text data are liable to reduce the transparency and accountability of these systems. Community members have already objected to algorithmic risk assessment systems because they have no way to contest inputs or outputs and have little visibility into how tools are constructed or different features are weighted \citep{Brown2019}. Text-processing systems are even more difficult to interpret than statistical classifiers: the volume of ongoing research on interpreting NLP models evidences that interpretability is an unsolved problem. Furthermore, although families likely have some knowledge of the values contained in structured data (e.g., if a caregiver has a criminal history), they typically have no knowledge of what contact notes contain nor the ability to contest them.

Information extraction systems focus on organizing information over direct decision-making, making them appear safer, and CPS agencies are more actively interested in deploying them; however, we do identify racial bias as a concern in this setting, and there are additional risks. Like with risk predictive models, deployment typically requires sharing data externally, and information extraction systems in particular tend to require much annotated data to achieve reasonable performance \citep{xia-van-durme-2021-moving}. Further, even with a hypothetical perfect model,  usefulness is entirely dependent on data quality. \citet{moon2023rethinking} demonstrate that notes often contain descriptions of \textit{perceived} risks that may not reflect reality, high turnover leads to inexperienced caseworkers, and caseworkers are incentivized to practice defensive decision-making over objective recording.
Our data reveals similar evidence that caseworkers write notes explicitly to document and justify decisions (Appendix~\ref{sec:decision_documentation}).
Word statistics (Section~\ref{sec:data_statistics}) also demonstrate that language can differ depending on the race of people involved.
A CPS worker may be able to distinguish reliable from unreliable information more easily when manually reading notes than when viewing outputs of information extraction systems, which reduce relevant context.
Finally, more investigation is needed to understand effect on clients. For example, DHS workers may find information extraction useful for preparing for court hearings, possibly even reducing administrative burden enough to decrease caseworker turnover. However, families and their advocates could be unfairly disadvantaged without access to similar tools and information.
Even if algorithmic racial bias could be mitigated, models can still perpetuate harms by retrieving biased information from underlying data and increasing the power imbalance between CPS agencies and affected families.

We focus on risk assessment and information extraction models, because they have clear use cases and are of active interest to CPS agencies, but recent advances in interactive generative models, like the highly publicized ChatGPT have resulted in much interest around using these models as well. Our work does include investigations of pre-trained language models: we examine RoBERTa for risk assessment (Appendix~\ref{sec:appendix_nlp_models}), the flairNLP NER models use XLM-R embeddings (Section~\ref{sec:ner_methods}, \citep{schweter2020flert, conneau-etal-2020-unsupervised}), and the neural coreference models use SpanBERT (Section~\ref{sec:ner_methods}, \citep{joshi2020spanbert}). Generative models may have the potential to be useful for tasks like summarization, court preparation, or perhaps even information extraction. However, current models are prone to hallucinating fictional information \citep{Ji2022}, severely imitating their usability in high-stakes domains. Popular models are additionally closed-sourced and require interfacing through APIs rather than running models locally, which precludes evaluating them over protected data. Our results suggest that should these models be considered for deployment, evaluating them for racial bias is essential.

\paragraph{Conclusions}
Studying the performance of NLP models in realistic high-stakes settings is extremely difficult due to concerns about data privacy and lack of transparency from many practitioners.
Nevertheless, advances in NLP model performance over benchmark datasets and interactive demos like ChatGPT have generated intense interest in deploying these types of models, which mandates understanding how they actually perform.
Our work aims to a provide a more realistic examination of algorithmic unfairness in NLP models than current research focused on synthetic benchmark data.
We do uncover some evidence of algorithmic racial bias in this setting, specifically in NER models and in documenting different language use.
Nevertheless, algorithmic unfairness is only one metric in this context, and more research is needed to uncover sociotechnical forces involved in deploying NLP models, including deeper engagement with stakeholders like affected families, caseworkers, and community advocates.

\paragraph{Limitations and Ethical 
Considerations}

As our work does not contribute to the development of deployable systems, misuse potential of this work is low. However, there is significant risk of our results being taken out of context or misinterpreted. We emphasize that we study specific models in specific contexts, and our results cannot be assumed to generalize to other scenarios without appropriate evaluation.
There are also inherent risks associated with working with such high-stakes data. We take numerous steps to mitigate these risks, including abiding by IRB and data sharing protocols, viewing anonymized versions of contact notes for data analyses as much as possible, and not providing any specific examples from notes in this paper. Given the interest in deploying NLP in high stakes settings, we believe the importance of providing visibility into model performance outweighs these risks. 

\begin{acks}
We gratefully thank the anonymous county Department of Human Services for providing feedback and data for this work and Lucille Njoo for providing feedback.
This work was supported by seed funding from the Block Center for Technology and Society at Carnegie Mellon University, Pittsburgh, PA.
A. Chouldechova has received research funding from the Allegheny County Department of Human services to support studies of racial bias in algorithm-assisted decision making.
A.F. acknowledges support from a Google PhD fellowship. A. Coston acknowledges support from the National Science Foundation Graduate Research Fellowship Program under Grant No. DGE1745016 and from a Meta PhD fellowship. Y.T. acknowledges support from NSF CAREER Grant No.~IIS2142739, the Alfred P.~Sloan Foundation Fellowship, and NSF grants No.~IIS2125201, IIS2203097, and IIS2040926.
\end{acks}

\balance
\bibliographystyle{ACM-Reference-Format}
\bibliography{refs}

\newpage

\appendix

\section{Alternative NLP Models for Risk Assessment}
\label{sec:appendix_nlp_models}

In the main paper, we focus on performance of a random forest model. Here, we additionally describe implementations and results for three other types of classification models: logistic regression, a convoluted neural network (CNN)-based neural classifier \citep{ji-etal-2021-medical}, and a RoBERTa-based neural classifier \citep{Liu2019RoBERTaAR}. In general, we compare training models with only structured features (structured) to training with only text features (text) to training with both structured features and text (hybrid). For the GatedCNN and RoBERTa models, we do not train structured-only versions, as these models were specifically designed for text processing.

\paragraph{GatedCNN} This model is a state-of-the-art classifier developed for assigning codes to medical notes \cite{ji-etal-2021-medical}. It uses a CNN-based architecture that involves injecting word embedding between layers and an LSTM-style gating mechanism. The original model additionally computed dot-product interactions between medical notes and codes, using word embeddings derived from code descriptions. In our hybrid model, we incorporated structured features by replacing the medical code representations with the 818-dimensional structured feature vectors.
For the GatedCNN model,  associated notes were truncated to the last (most recent) 3,000 tokens. The model embeddings were initialized with 100-dimensional word embeddings trained over the full data set of 3.1M contact notes (excluding the test set) using skip-gram Word2Vec with a context window of 5. We used the same kernel, filter sizes, and hidden layer sizes as the original model \cite{ji-etal-2021-medical}. The model was trained with a learning rate of 1e-03 for up to 20 epochs, with early stopping if development set performance did not increase for 3 epochs. Hyper-parameters were selected based on development set performance after 5 epochs of training.

\paragraph{RoBERTa} We trained a RoBERTa-based classifier \cite{Liu2019RoBERTaAR}, where classification decisions were made using the final CLS representation.
Popular variants of the high-performing transformer architecture (including RoBERTa) only support inputs up to 512 tokens \cite{Liu2019RoBERTaAR}. Thus, we concatenated all associated notes and truncated them to the last (most recent) 512 tokens. In early experiments, we found that truncation outperformed alternative approaches to handling long inputs to a transformer, such as selecting input sentences using scoring functions or hierarchical models \cite{Pappagari2019HierarchicalTF}, which likely require larger training corpora. To incorporate structured features, we concatenated them to the CLS representation and passed the concatenated vector through a fully-connected linear layer, followed by a soft-max layer.

Prior to training the model, we conducted additional masked-language-model pretraining over the full data set of 3.1M contact notes (excluding test data) for 1 epoch, which prior work has shown improves performance on domain-specific data \cite{gururangan-etal-2020-dont}. We fine-tuned the model for classification using a learning rate of 1e-05 and weight decay of 0.01 for up to 30 epochs, where training was stopped early if development set performance did not increase for 3 epochs. As above, hyper-parameters were selected based on development set performance after 5 epochs of training.

\begin{figure}
    \centering
    \begin{minipage}[c]{0.4\textwidth}
    \includegraphics[width=\textwidth]{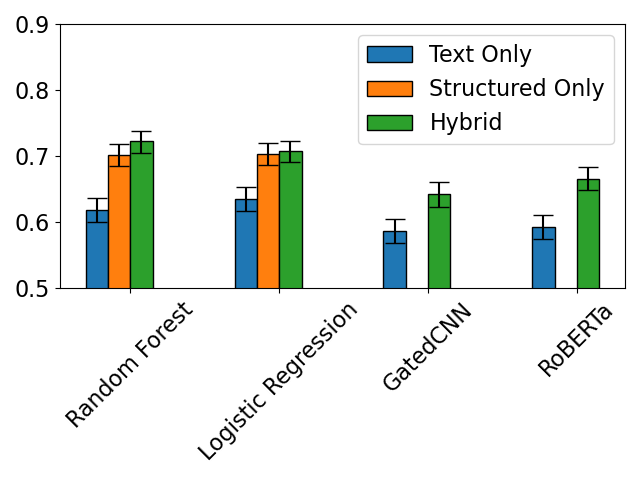}
\end{minipage}
\begin{minipage}[c]{0.55\textwidth}
    \begin{tabular}{cccc}
        \hline
      \multicolumn{2}{c}{Common Words} & \multicolumn{2}{c}{Highly Weighted Words}  \\
      \hline
      \hline
    Caseworker & Cas, ework, er & twin & tw, in  \\
    asked & ask, ed  & lice & l, ice \\
    Grandmother & Grand, mother & decision & dec, ision\\
    Maternal & M, aternal & concern & con, cern\\
    visit  & vis, it & custody & c, ust, ody \\
    concerns &con, cern, s & probation & pro, b, ation\\
    Youth & Y, outh & unexcused & un, exc, used\\
    Intake & Int, ake & paramour & param, our\\
    Families &  F, am, ilies & spank & sp, ank \\
    Paternal &  P, aternal & furniture & f, urn, iture \\
    denied & den, ied \\
    \end{tabular}
\end{minipage}
    \caption{Top: AUC scores of model variants. Metrics only include test data with text. 90\% confidence intervals are computed using bootstrap sampling over the test set.
    Bottom: Words that RoBERTa tokenizer splits into to word pieces, includes the most common words in the training corpus that are split and the words assigned the highest or lowest weights by text-only logistic regression classifiers.}
    \label{fig:risk_auc_ci}
    \Description{Bar chart with 3 series: one for text-only models, one for structured models, and one for hybrid models. The x-axis show 4 models: random forest, logistic regression, GatedCNN, and RoBERTA. The bars are highest for the hybrid series across all models. The bars for the random forest model are the highest overall. The bars for the GatedCNN and RoBERTa models are much lower than for the random forest and logistic regression models.}
\end{figure}

RoBERTa uses Byte-Pair Encoding (BPE) to enable handling large vocabulary, which divides out-of-vocabulary words into sub-pieces in order to derive components that are part of the vocabulary \cite{Liu2019RoBERTaAR,Radford2019,sennrich-etal-2016-neural}. In \Cref{fig:risk_auc_ci}, we show some of the most common words in our corpora that were not included in the model vocabulary and were sub-divided by the tokenizer.

\paragraph{Random Forest and Logistic Regression} For the text models, 
we extracted TF-IDF-weighted bag-of-words features using a 10,000 word vocabulary.
For the hybrid models, early experiments showed that concatenating the full 10,000 text features with the 818 structured features caused the model to ignore the structured features. Instead, we first trained a text model using a logistic regression classifier. We then took the 500 words with the highest learned coefficients and the 500 words with the lowest (most negative) coefficients and constructed TF-IDF features from this 1,000 word vocabulary. We then concatenated these features with the 818 structured features, constructing 1,818-dimensional feature vectors. All random forest classifiers used 500 trees and all logistic regression classifiers used L2 regularization.

\paragraph{Results} \Cref{fig:risk_auc_ci} reports AUC scores for the different models. Across all model variants, the text models performed worse (random forest: 61.89, logistic regression: 63.50, GatedCNN: 58.66, RoBERTa: 59.19) than the structured (random forest: 70.19, logistic regression: 70.29) or hybrid (random forest: 72.21, logistic regression: 70.74, GatedCNN: 64.24, RoBERTa: 66.60) models. We also observed that the GatedCNN and RoBERTa models performed worse than the bag-of-words statistical classifiers. Based on the model performance results in Figure 1, in the main paper, we analyzed the performance of the random forest structured and hybrid models.


\section{Additional Model Metrics}
\label{sec:additional_model_metrics}

\begin{table}[h!]
    \centering
    \begin{tabular}{ccccccc}
    \hline
      & \multicolumn{2}{c}{All (14,417)} & \multicolumn{2}{c}{Black (6,841)}  & \multicolumn{2}{c}{White (5,763)} \\
      \hline
      \hline
       & Struct. & Hybrid & Struct. & Hybrid & Struct. & Hybrid \\
       \hline
AUC & 75.77 & 76.27* & 74.70 & 75.69* & 74.96 & 74.96 \\
TPR & 56.04 & 56.40* & 56.81 & 57.46* & 53.94* & 53.46 \\
FPR & 19.59 & 19.53* & 20.79 & 20.36* & 19.69* & 20.01 \\
FNR & 43.96 & 43.60* & 43.19 & 42.54* & 46.06* & 46.54 \\
Precision & 32.47 & 32.67* & 37.02 & 37.77* & 28.18* & 27.67 \\
F1 & 41.12 & 41.38* & 44.83 & 45.58* & 37.02* & 36.47 \\
Accuracy & 76.91 & 77.01* & 75.25 & 75.71* & 77.00* & 76.66 \\
\bottomrule
    \end{tabular}
    \caption{Metrics for risk prediction task with different variants of the random forest model over all test data. Where the difference between the hybrid and structured models is significant ($p < 0.05$) the better-performing value is starred.}
    \label{tab:accuracy_all_data}
\end{table}

\begin{table}[h!]
    \centering
    \begin{tabular}{ccccccc}
    \hline
      & \multicolumn{2}{c}{All (4,133)} & \multicolumn{2}{c}{Black (1,880)}  & \multicolumn{2}{c}{White (1,894)} \\
      \hline
      \hline
       & Struct. & Hybrid & Struct. & Hybrid & Struct. & Hybrid \\
       \hline
AUC & 69.83 & 71.88* & 68.26 & 70.72* & 70.09 & 71.77* \\
TPR & 59.49 & 70.20* & 59.45 & 71.76* & 58.53 & 67.32* \\
FPR & 31.78* & 40.34 & 33.42* & 43.55 & 30.70* & 37.37 \\
FNR & 40.51 & 29.80* & 40.55 & 28.24* & 41.47 & 32.68* \\
Precision & 31.37* & 29.82 & 36.42* & 34.68 & 26.56* & 25.47 \\
F1 & 41.07 & 41.85* & 45.17 & 46.75* & 36.54 & 36.95* \\
Accuracy & 66.51* & 61.72 & 64.84* & 60.18 & 67.58* & 63.38 \\
\bottomrule
    \end{tabular}
    \caption{Metrics for risk prediction task with different variants of the random forest model over test data that contains text. Where the difference between the hybrid and structured models is significant ($p < 0.05$) the better-performing value is starred.}
    \label{tab:accuracy_text_only}
\end{table}


\section{Decision Documentation Experimental Results}
\label{sec:decision_documentation}

We investigated whether text data is liable to increase overfitting to human decisions by examining how predictive contact notes are of near-term decisions. Specifically, given a referral-child observation that was screened in, we trained a model to predict if the referral will be accepted for services, and we separately trained a model to predict if the child will be placed (removed from home) within 2 years of the referral.
For each task, we trained structured and hybrid models, where we incorporated notes associated with the current referral into the hybrid model. Thus, while our primary hybrid model incorporated notes that precede screening decisions, these models incorporated notes typically generated during the investigation phase, after a screening decision has already been made.
As most screened-in referrals have associated notes generated during investigation, in this data set, 28,340/28,769 training observations had notes and all 14,417 test observations had notes.
 
\Cref{tab:performance_supervised_dec} reports the AUC scores of the random forest structured and hybrid models. The hybrid models showed strong improvements over the structured models. \Cref{tab:performance_supervised_dec} also reports the top-weighted words by text logistic regression models over the same data. 
The model assigns the highest weights to "decis" (lemmatized decision), "hear," "servic" (lemmatized service), and "crisi" (lemmatized crisis).
These results provide evidence that notes often contain documentation and justification of decisions.

\begin{table}[h!]
    \begin{tabular}{c|c|c|}
         Model & Service & Placement \\
        \hline
        Structured AUC & 77.2 & 75.9 \\
        Hybrid AUC & 86.1 & 80.5 \\
        \hline
        \hline
       Highest-weighted & crisi, servic, hear, & foster, placement, hear, \\
        Words &  decis, group &   author, physic \\
    \end{tabular}
    \caption{\emph{Top:} AUC for random forest models with and without text features from notes directly associated with referrals over two nearer-term decisions: whether or not the referral will be accepted for services and out-of-home placement. \emph{Bottom:} words assigned the highest weights by a text logistic regression classifier. Notes often document and explicitly justify decisions.}
    \label{tab:performance_supervised_dec}
\end{table}

\end{document}